\documentclass{bmvc2k}
\usepackage{graphicx}
\usepackage{cleveref}
\usepackage{amsmath}
\usepackage{amssymb}
\usepackage{booktabs}
\usepackage{color,soul}
\usepackage{caption}
\usepackage{svg}
\usepackage{pgfplots}
\pgfplotsset{compat=1.17}
\usepackage{amssymb} 
\usepackage{relsize}
\usepackage{enumitem}
\usepackage{enumitem}
\usepackage{hyperref}
\usepackage[none]{hyphenat}
\usepackage{amsmath}
\usepackage{siunitx}
\crefname{figure}{Fig.}{Figs.}
\crefname{table}{Table}{Tables.}

\title{G3FA: Geometry-guided GAN for Face Animation}

\addauthor{Alireza Javanmardi}{alireza.javanmardi@dfki.de}{1}
\addauthor{Alain Pagani}{alain.pagani@dfki.de}{1}
\addauthor{Didier Stricker}{didier.stricker@dfki.de}{1}

\addinstitution{
 German Research Center for Artificial\\
 Intelligence (DFKI)\\
 Kaiserslautern, Germany
}

\runninghead{a.javanmardi, a.pagani, d.stricker}{G3FA}

\begin{document}

\maketitle

\begin{abstract}
Animating human face images aims to synthesize a desired source identity in a natural-looking way mimicking a driving video's facial movements.
In this context, Generative Adversarial Networks have demonstrated remarkable potential in real-time face reenactment using a single source image, yet are constrained by limited geometry consistency compared to graphic-based approaches. In this paper, we introduce Geometry-guided GAN for Face Animation (G3FA) to tackle this limitation. Our novel approach empowers the face animation model to incorporate 3D information using only 2D images, improving the image generation capabilities of the talking head synthesis model. We integrate inverse rendering techniques to extract 3D facial geometry properties, improving the feedback loop to the generator through a weighted average ensemble of discriminators. In our face reenactment model, we leverage 2D motion warping to capture motion dynamics along with orthogonal ray sampling and volume rendering techniques to produce the ultimate visual output. To evaluate the performance of our G3FA, we conducted comprehensive experiments using various evaluation protocols on VoxCeleb2 and TalkingHead benchmarks to demonstrate the effectiveness of our proposed framework compared to the state-of-the-art real-time face animation methods. Our code is available at \href{https://github.com/dfki-av/G3FA}{github.com/dfki-av/G3FA}.
\end{abstract}

\section{Introduction}
\label{sec:intro}
Talking head generation involves the task of re-rendering a source face image with a new pose and expression, often controlled by either the same individual or a different one.
This technology finds particular application in video conferencing tools by enabling significant bandwidth reduction through keypoints transmission or by displaying alternative appearance representations. This not only reduces transmission costs but also mitigates potential biases in scenarios like job interviews while preserving a sense of presence for participants. \\ 
While this challenge was initially tackled by computer graphics researchers, new advances in deep generative modeling led to significant progress in this field  \cite{siarohin2019first,zakharov2020fast,wang2021one,pang2023dpe,drobyshev2022megaportraits}.
\newline Approaches centered around graphics-based head synthesis demonstrate impressive geometric capabilities as they aim to fully reconstruct and animate human heads \cite{Thies_2016_CVPR}. Typically, these approaches leverage parametric models \cite{blanz1999morphable,li2017learning} to reconstruct the geometry of the human's head, subsequently mapping the source identity's texture to render the animation \cite{Thies_2016_CVPR}. Recent advances in this domain, particularly by the introduction of 3D Gaussian Splitting technique \cite{kerbl20233d} are moving towards increased photorealism level and also rendering speed \cite{xu2023gaussianheadavatar}. Nonetheless, this approach necessitates multiple images captured from various viewpoints of the subject's head and entails some processing time for head reconstruction.
\newline 
In parallel, researchers have been harnessing deep generative models to enhance face reenactment techniques. This is typically achieved by extracting features from a desired source image and warping them based on a driving video \cite{siarohin2019first,zeng2022fnevr,mallya2022implicit} or utilizing a pre-trained StyleGAN2 model \cite{karras2020analyzing} and navigation in the latent space \cite{wang2022latent,yin2022styleheat}. Compared to graphics-based approaches, these methods demonstrate higher generalization capabilities, having been trained on thousands of video clips and capable of animating diverse human face images. They can start rendering instantaneously using a single shot of the source identity, producing photorealistic outcomes including fine details and accessories like glasses. However, since these methods are trained on 2D data, they are generally less robust in terms of geometry reconstruction. This leads to poor reconstruction quality in the case of head rotations far from the frontal view - typically when the driving source turns the head left or right.\newline To overcome these limitations, this paper introduces a novel face animation method based on Generative Adversarial Networks (GAN) \cite{goodfellow2020generative} that incorporates 3D supervision to enhance the generator's awareness of the true distribution of the human's head while synthesizing novel head poses. To address this challenge, an inverse rendering approach is employed to extract geometry properties including depth and normal maps from RGB frames. Our method uses a weighted combination of discriminators, to inject this information into the generator as a core module for image synthesis. The proposed framework can be easily applied to most adversarial-based face reenactment models, leveraging off-the-shelf pre-trained inverse rendering models.
\newline The contributions of this paper can be summarized as follows: 
\begin{itemize}[itemsep=0pt]
\item We propose implicit 3D supervision leveraging prior human head information for face reenactment models to enhance geometry consistency without affecting inference time.
\item We introduce a novel integration of 3D properties into a GAN framework for face animation tasks, without imposing significant computational overhead during training.  
\item  We conduct extensive experiments across various scenarios on benchmark datasets, demonstrating a superior realism compared to recent state-of-the-art approaches, both quantitatively and qualitatively, particularly in the case of extreme head pose variations.\end{itemize}

\begin{figure*}
  \centering
  \includegraphics[width=1\columnwidth]{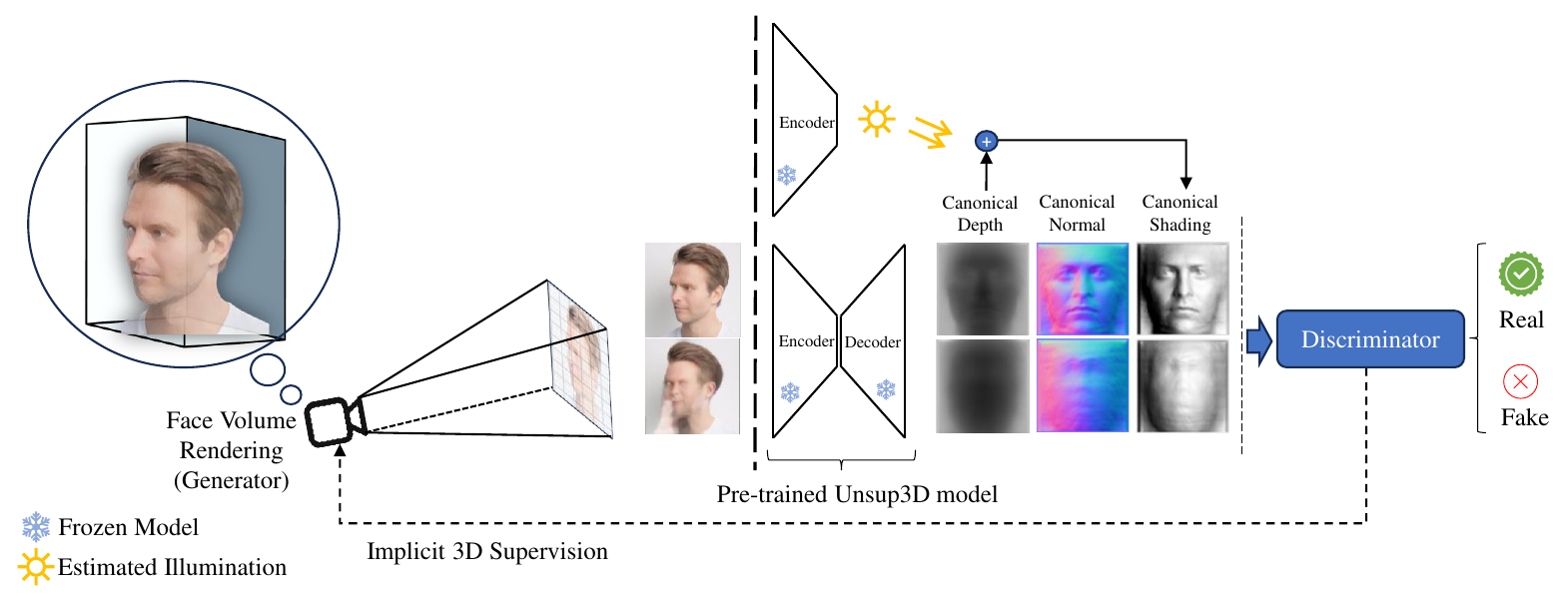}
  \captionsetup{skip=5pt} 
  \caption{Implicit 3D supervision: This figure shows how an inverse rendering module can guide the generator to generate more geometry-consistent output. We utilized canonical shading here to better visualize the differences between two cases and it is not used in the model's pipeline.}
  \label{fig:figure1}
\end{figure*}
\section{Related Works}

\subsection{Face Reenactment}

Recent reconstruction-based methodologies typically employ the FLAME 3D face model \cite{li2017learning} and rendering algorithms, such as rasterization, to achieve real-time reenactment. These approaches involve reconstructing the avatar's head and subsequently animating it by manipulating the morphable model's parameters. However, these techniques exhibit limitations in rendering fine details and entail significant training time when reconstructing heads from 2D-captured data. \cite{Thies_2016_CVPR, Grassal_2022_CVPR}. Generative models, including GANs and Diffusion models \cite{ho2020denoising}, have demonstrated significant success in this field. In the real-time domain,
GAN-based techniques initiate by estimating 2D or 3D keypoints and deriving the optical
flow of these points between the source and driving frames. Following this, they deform the features of the source image and utilize a generator model to produce the animated output \cite{siarohin2019first,wang2021one,zakharov2019few,hong2022depth}. Expanding on the concept of GAN inversion  \cite{shen2021closed}, another research direction, aims to animate face images based on latent space navigation of StyleGAN model \cite{wang2022latent,bai2023high,yin2022styleheat}. 
On the other hand, Diffusion models provide face animation approaches that can even work with other modalities like audio information \cite{tian2024emo,shen2023difftalk,xu2024vasa}. These models are typically used for offline tasks where rendering time is not crucial. 
Additionally, some researchers have explored the application of neural rendering techniques within an adversarial framework, capitalizing on their ability to generate photorealistic results \cite{ren2021pirenderer,yin2023nerfinvertor,zeng2022fnevr}.

\subsection{Generative Models}

Generative adversarial networks (GANs) \cite{goodfellow2020generative} serve as the primary engine for generating high-fidelity samples in the domain of talking head synthesis. Researchers have endeavored to expedite convergence and enhance the stability of GAN models by optimizing their architectures \cite{isola2017image,mirza2014conditional} or loss functions \cite{arjovsky2017wasserstein,gulrajani2017improved}. In this regard, certain works have explored the use of two discriminators, whereby forward and backward KL divergence is computed, resulting in improved performance compared to the standard framework \cite{nguyen2017dual}. Additionally, the combination of pre-trained discriminators with varying architectures has been employed to enhance the quality of the generated images \cite{kumari2022ensembling,mordido2018dropout}. Some studies have also tried to guide the generator through more effective supervision by incorporating geometry information during the model's training process \cite{shi2022improving, hong2022depth, wang2021safa}.
 
\begin{figure*}
  \centering
  \includegraphics[width=1\columnwidth]{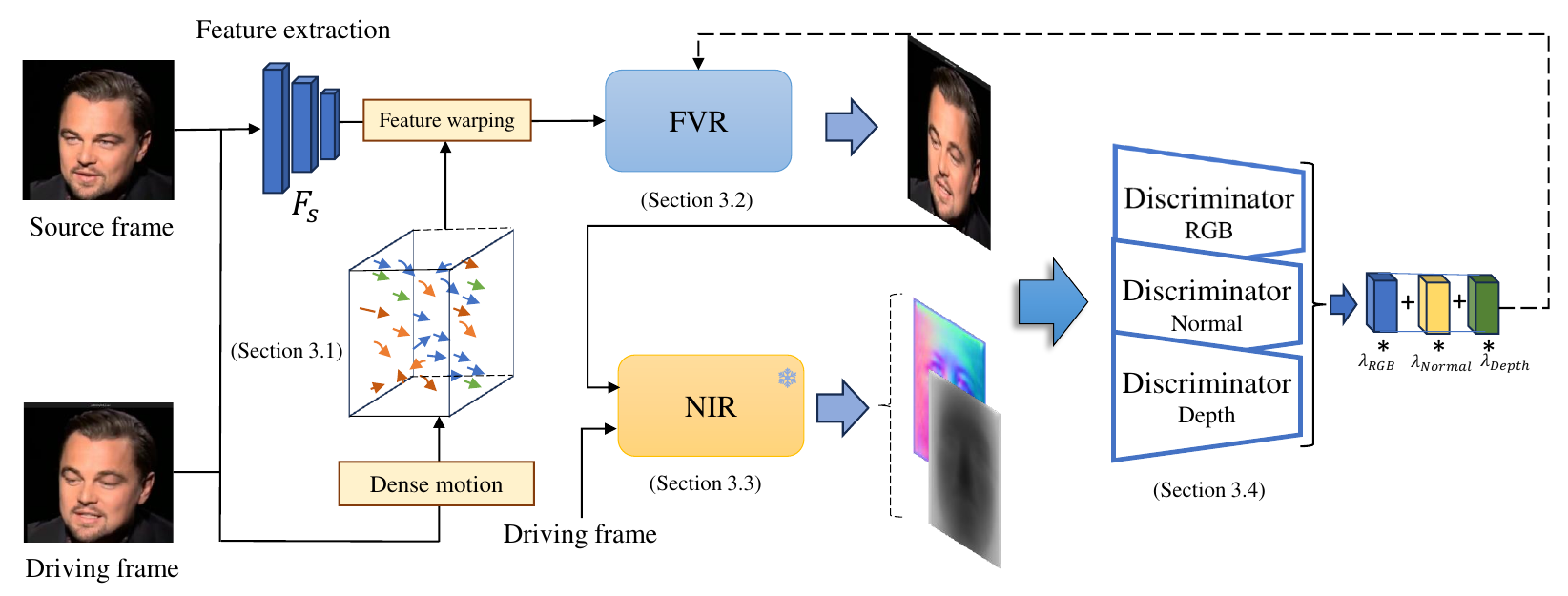}
  \captionsetup{skip=5pt}
  \caption{Face animation pipeline: Capturing facial expression and pose based on keypoints, followed by an implicit 3D supervision using inverse rendering and an ensemble of discriminators. NIR stands for Neural Inverse Rendering which is a pre-trained model and FVR is our Face Volume Rendering module.}
  \label{fig:figure2}
\end{figure*}


\section{Method}
The global architecture of our method is shown in \cref{fig:figure2}. We leverage the face volume rendering (FVR) technique \cite{max1995optical} in conjunction with an adversarial training procedure. To empower the model in generating photorealistic outcomes mainly based on RGB training data, we integrated a neural inverse rendering module and two additional discriminators to process the extracted 3D attributes. Our objective is to synthesize images with the appearance of a given source image  $I_{S}$ while incorporating the pose and facial expressions from a set of driving frames $\left\{I_{D_{1}}, I_{D_{2}},..., I_{D_{n}}\right\}$, which could be sourced from videos or live webcam streams. In the subsequent sections, we provide comprehensive information on each individual module.

\subsection{2D Motion Estimation}
We start by extracting features from the source image $f_{s}$ using a dedicated self-supervised feature extraction module $F_{s}$. To capture the facial dynamics of the driving frames, we employ a motion estimation module. Within this module, we detect and extract a set of $k=15$ 2D landmarks from both the source and driving frames, leading to pairs  $\{q_{s,k},q_{d,k}\} \in \mathbb{R}^{2}$. Subsequently, to better emulate facial changes, we consider the neighboring points as a $2\times2$ Jacobian matrix $\{J_{s,k},J_{d,k}\} \in \mathbb{R}^{2\times2}$ for each landmark. Using first-order Taylor expansion, we can obtain an affine approximation of motion field modeled by  $\tau_{S\leftarrow D}$ based on backward optical flow as in \cite{siarohin2019first}: \\
\vspace*{-4mm}
\begin{equation}
\tau_{S\leftarrow D,k}(z) \approx q_{s,k}+J_{s,k} {J_{d,k}}^{-1}(z-q_{d,k}).
  \label{eq:important}
\end{equation}
Although some studies have advocated for the utilization of 3D keypoints and the elimination of the Jacobian matrix \cite{wang2021one}, we did not observe significant differences. Moreover, adopting 3D keypoints introduces a substantial number of parameters to the model. Leveraging a dense motion field for each keypoint \cite{siarohin2019first}, we combine them with a weight factor as a mask, ranging from 0 to k $ \left\{M_{0},M_{0} , ...,M_{K} \right\} $ , where the initial one represents the background. The dense motion estimation module also produces an occlusion map $O$ to generate the occluded parts using Hadamard product while synthesizing novel poses:\\
\vspace*{-3mm}
\begin{equation}
\hat{\tau}_{S\leftarrow D}(z) =M_{0}z+\sum_{k=1}^K M_{k} {\tau}_{S\leftarrow D,k}(z),
  \label{eq:important}
\end{equation}
\vspace*{-3mm}
\begin{equation}
F_{w}=O \odot f_{w}(f_{s},\hat{\tau}_{S\leftarrow D,k}).
  \label{eq:important}
\end{equation}

Finally, in this part of the framework, the extracted features from the source image are warped using information obtained from the dense motion estimation module. This allows the warped features to be utilized by the face volume rendering component.

\subsection{Rendering}

To render the final animated face results, we use the face volume rendering architecture proposed by \cite{zeng2022fnevr}. This architecture, compared to other approaches utilizing a single U-Net model \cite{ronneberger2015u} as the generator, offers a reduced parameter count and the ability to generate photorealistic outcomes through orthogonal ray sampling and volume rendering. It is important to note that our proposed framework seamlessly integrates with other architectures that follow an adversarial training procedure. \\ To employ the face volume rendering module, which consists of several parts, we initially feed the warped features $F_{w}$ obtained from the previous stage into two networks: the 3D shape extractor $\phi_{\sigma}$  and the color extractor $\phi_{color}$ similar to \cite{zeng2022fnevr}. These networks serve to both separate these two types of information from the features and elevate them to a higher-dimensional space. In order to calculate the sampled color and density, we leverage the proposed orthogonal ray-sampling module, which employs $f_{\theta}$; a multi-layer perceptron (MLP) to estimate the color field $F_{color}$ and voxel probability $F_{\sigma}$ in an adaptive manner \cite{zeng2022fnevr}, instead of adhering to the ray-sampling strategy proposed in NeRF \cite{mildenhall2021nerf}:
\\
\vspace*{-3mm}
\begin{equation}
F_{\sigma}=\phi_{\sigma}(F_{w}),
  \label{eq:important}
\end{equation}
\begin{equation}
F_{color}=\phi_{color}(F_{w}),
\label{eq:important}
\end{equation}
\begin{equation}
p_{\sigma},p_{color}=f_{\theta}(F_{\sigma},F_{color}).
  \label{eq:important}
\end{equation}
\vspace*{-3mm}
\\
We confine the camera pose to the frontal view and employ an adaptive ray sampling method, which significantly enhances the rendering speed. In our framework, we did not employ a 3D face reconstruction component to supervise the rendering model. However, by employing geometry guidance based on 3D geometric properties and an ensemble of discriminators, we can provide implicit supervision to the rendering model. To aggregate the color $p_{color}$ and the density  $p_{\sigma}$ obtained from the adaptive ray sampling segment, we utilize a volume rendering algorithm, albeit distinct from \cite{max1995optical}:
\\
\vspace*{-3mm}
\begin{equation}
F_{r,i}=\sum_{j=1}^D \tau_{j}(1-exp(-p_{\sigma,i,j}))p_{color,i,j},
  \label{eq:important}
\end{equation}
\vspace*{-3mm}
\begin{equation}
\tau_{j}=exp(\sum_{k=1}^{j-1}-p_{\sigma,i,k}),
  \label{eq:important}
\end{equation}
where $\tau_{j}$ is the transmittance term. This algorithm enables the derivation of the final RGB value. Furthermore, we have observed that concatenating the extracted features from the source image $f_{s}$ with the output of the volume rendering module $F_{r}$ and passing them through a shallow decoder with SPADE layers improves the fidelity of the results, as suggested in \cite{zeng2022fnevr}. A detailed comparison of computational costs is provided in the supplementary material.
\subsection{Neural Inverse Rendering}
As depicted in \cref{fig:figure1}, we utilized a well-established Unsup3D model \cite{wu2020unsupervised} to obtain 3D geometric properties, such as canonical depth and normal map, from a single RGB image. These properties are subsequently employed to enhance the quality of the rendered results. By regarding the RGB image $I$ as a function $\Omega\rightarrow \mathbb{R}^{3}$ that maps coordinate $\{0,...,W-1\} \times \{0,...,H-1\}$ in a 2D space to tensor in $\mathbb{R}^{3\times W \times H}$, and under the assumption that the image is centered and relatively symmetric, we adopt an autoencoder architecture model, as in \cite{wu2020unsupervised}, to obtain 3D properties, including depth map $d$, global light direction $l$, viewpoint $w$, and albedo image $a$, using an analysis-by-synthesis approach making $\hat{I} \approx I$:
\vspace*{-2mm}
\begin{equation}
\hat{I}=\prod (\psi(a,d,l),d,w),
  \label{eq:important}
\end{equation}
Here, $\psi$ synthesizes a version of the object from a canonical viewpoint, then the reprojection function $\prod$ tries to simulate the viewpoint effect to the model using shaded canonical image $\psi(a,d,l)$ and also depth map $d$. However, in our framework, we utilize a pre-trained Unsup3D model trained on the CelebA dataset \cite{liu2015deep}, extracting only depth information, from which we subsequently derive the normal map.
\subsection{Ensemble of Discriminators}
To incorporate the extracted 3D information, obtained through prior knowledge of the human face, into our talking head synthesis pipeline, we introduce two additional discriminators. For this purpose, we feed each discriminator with the RGB image, canonical depth, and normal map, respectively.
All of these discriminators employ a multi-scale architecture \cite{Wang_2018_CVPR} with spectral normalization \cite{miyato2018spectral}. To ensure the convergence of the model is unaffected, we assign different weights $\lambda_{i}$ to each discriminator \cite{667881} and aggregate them based on the min-max formulation to obtain the final result:
\vspace*{-2.5mm}
\begin{equation}
\begin{aligned}
\mathcal{L}_{\text{GAN}}(G, D_{total}) = \mathbb{E}_{\mathbf{x} \sim p_{\text{real}}(\mathbf{x})} [\log D_{total}(\mathbf{x})] 
+ \mathbb{E}_{\mathbf{x}^\prime \sim p_{\mathbf{fake}}(\mathbf{X}^\prime)} [\log(1 - D_{total}(\mathbf{x}^\prime))],
\end{aligned}
\end{equation}
\vspace*{-3mm}
\begin{equation}
\begin{aligned}
D_{total} = \sum_{i\in S} \lambda_{i}D_{i}(x_{i}),
\end{aligned}
\end{equation}

where $S =\{ \text{RGB, depth, normal}\}$  and we assume $\sum_{i} \lambda_{i}=1$. By incorporating the 3D face geometry as additional supervision, the generative model can better estimate the true data distribution. The photorealistic results generated by the Unsup3D model produce meaningful depth and normals, thereby reducing the adversarial loss, rendering them close to real images. This integration of 3D information into a 2D GAN pipeline, in contrast to \cite{shi2022improving} and \cite{hong2022depth}, preserves the inference time efficiency, and when combined with more powerful inverse rendering techniques, can yield even more effective outcomes.

\section{Experiment}
The G3FA framework is devised to operate in a fully self-supervised manner during the training process. It randomly extracts two frames from a video: the first frame serves as the source, while the second frame acts as the driving frame. The framework then animates the source image based on the changes exhibited by the driving frame, while simultaneously incorporating the ground-truth animated frame for comparison. Adversarial and perceptual losses \cite{johnson2016perceptual} are employed to ensure the photorealism of the generated samples. The adversarial loss encompasses the combined predictions of all discriminators, while the perceptual loss \cite{johnson2016perceptual} leverages a pre-trained VGG-19 model \cite{simonyan2014very} trained on ImageNet \cite{russakovsky2015imagenet} to extract features. Additionally, to ensure the consistency of detected keypoints in the face, the framework incorporates an equivariance loss \cite{siarohin2019first}. The keypoint extraction module extracts keypoints and their jacobians, all of which are attained in a fully self-supervised manner:
\vspace*{-2mm}
\begin{equation}
\begin{aligned}
\mathcal{L}_{\text{total}} = \mathcal{L}_{P}(x, x^\prime) + 
\mathcal{L}_{\text{GAN}}(\{x_i, x^\prime_{i}\}_{i \in S}) + \mathcal{L}_{E}(\{x_{d,k}\})
\end{aligned}
\end{equation}

\begin{table*}[t]
\centering
\captionsetup{skip=5pt} 
\caption{Comparison with prior works on same-identity reconstruction on VoxCeleb2~\cite{chung18b_interspeech}. Bold values indicate the best performance, while underlined values represent the second-best.}
\label{Table:comparison1}
\begin{tabular}{lcccccc}
\toprule
\textbf{Method} & \textbf{L1 $\downarrow$} & \textbf{LPIPS $\downarrow$} & \textbf{PSNR $\uparrow$} & \textbf{SSIM $\uparrow$} & \textbf{FID $\downarrow$} & \textbf{AKD $\downarrow$} \\
\midrule
FOMM \textit{\tiny (NeurIPS 2019)} & 12.31 & 0.109 & 23.52 & 0.71 & 24.59 & 1.89 \\
Face vid2vid \textit{\tiny (CVPR 2021)} & \underline{11.13} & 0.125 & 24.15 & 0.84 & 21.78 & 1.72 \\
DaGAN \textit{\tiny (CVPR 2022)} & 11.22 & 0.117 & \textbf{25.64} & \underline{0.88} & 22.83 & \underline{0.91} \\
FNeVR \textit{\tiny (NeurIPS 2022)} & 11.16 & \underline{0.094} & 24.18 & 0.77 & 21.11 & 0.95 \\
LIA \textit{\tiny (ICLR 2022)}& 12.02 & 0.106 &  \underline{25.57}  & 0.84 & \textbf{16.47} & 1.12 \\
HyperReenact \textit{\tiny (ICCV 2023)} & 13.42 & 0.111 &  22.51  & 0.69 & 28.87 & 1.28 \\
\midrule
G3FA (ours) & \textbf{10.87} & \textbf{0.081} & 24.51 & \textbf{0.91} & \underline{18.79} & \textbf{0.80} \\
\bottomrule
\end{tabular}
\end{table*}

\begin{table}[t]
\captionsetup{skip=5pt}
\centering
\caption{Quantitative comparison of cross-identity reenactment on VoxCeleb2\cite{chung18b_interspeech} and TK\cite{wang2021one}. Bold values indicate the best performance, while underlined values represent the second-best.}
\label{Table:comparison2}
\begin{tabular}{llcccc}
\hline
\multicolumn{1}{c}{\textbf{}} & \multicolumn{2}{c}{\textbf{VoxCeleb2}} & \multicolumn{2}{c}{\textbf{TK}} \\
\hline
\textbf{Method} & \textbf{FID} $\downarrow$ & \textbf{CSIM} $\uparrow$ & \textbf{FID} $\downarrow$ & \textbf{CSIM} $\uparrow$ \\
\hline
FOMM & 142.18 & 0.5219 & 130.78 & 0.5402 \\
Face vid2vid & 139.74 & 0.5971 & 121.44 & 0.6114 \\
DaGAN & 130.77 & 0.6021 & \underline{120.94} & 0.6264 \\
FNeVR & 132.36 & 0.5408 & 122.47 & 0.6021 \\
LIA & \textbf{122.26} & \underline{0.6078} & \textbf{118.59} & \underline{0.6338} \\
HyperReenact & 152.94 & 0.5144 & 147.63 & 0.4923 \\
\hline
G3FA (ours) & \underline{127.12} & \textbf{0.6274} & 122.83 & \textbf{0.6455} \\
\hline
\end{tabular}
\label{tab:comparison2}
\end{table}
\vspace*{-5mm}

\subsection{Implementation Details}
{\bf Datasets:} Our experiments were conducted using two well-known datasets, VoxCeleb2 \cite{chung18b_interspeech} and TK (TalkingHead-1KH) \cite{wang2021one}. VoxCeleb2 \cite{chung18b_interspeech} comprises over 1 million videos encompassing approximately 6K distinct identities, while TK offers about 180K high-quality samples. As part of the preprocessing stage, we traced and cropped face images from the videos, resizing them to a standardized dimension of $256\times256$, as in \cite{siarohin2019first}. This preprocessing methodology holds particular significance for our neural inverse rendering module, which was also trained on centered images, ensuring the acquisition of meaningful geometric properties.

\subsection{Training details}
Our G3FA model was trained on the pre-defined training set of VoxCeleb2 \cite{chung18b_interspeech}. Moreover, randomly selected 90\% of the videos from the TK dataset were included for training. For network optimization, we employed the Adam optimizer \cite{kingma2017adam} with the same parameters across all modules $\eta=2\times10^{-4}$, $\beta_{1}=0.5$ and $\beta_{2}=0.9$. After conducting several experiments, we determined the weights of the three discriminators as follows: $\lambda_{RGB}=50\%$ , and $\lambda_{depth}=\lambda_{normal}=25\%$ for depth and normal. These weights remained fixed throughout the training process. Ablation studies investigating various values of $\lambda$ are detailed in the appendix.

\subsection{Evaluation Metrics}
To gauge the reconstruction quality, we employed the L1 loss, Learned Perceptual Image Patch Similarity (LPIPS), Structural Similarity Index Metric (SSIM), Peak Signal-to-Noise Ratio (PSNR), Average Keypoint Distance (AKD) \cite{JMLR:v12:gashler11a} using MTCNN \cite{zhang2016joint} and to evaluate the identity preservation, we utilized cosine similarity metric (CSIM). Additionally, to compare the distributions and assess the quality and diversity of real and generated samples, we utilized the Fréchet Inception Distance (FID) metric.

\begin{figure*}[!t]
  \centering\includegraphics[width=0.95\columnwidth]{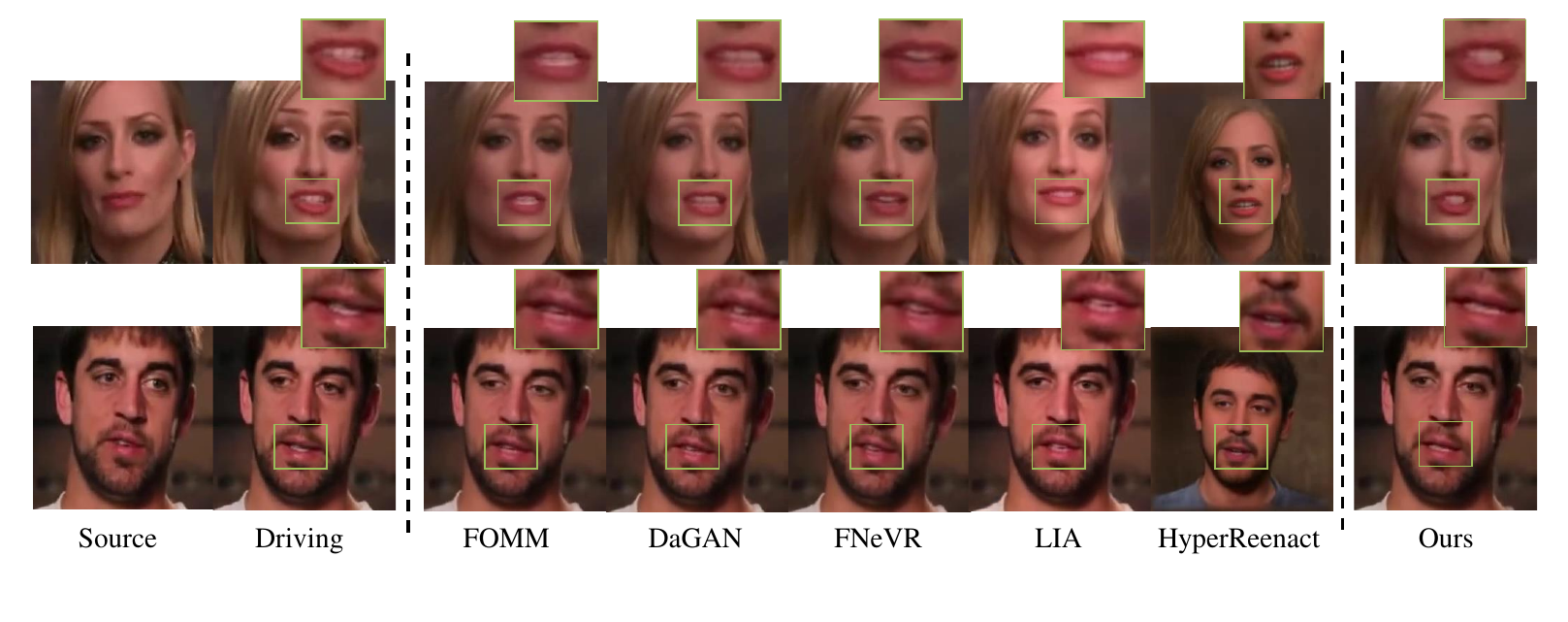}
  \caption{Same-identity reconstruction: Our method exhibits superior performance in terms of both photorealism image generation and precise synthesis of fine details on VoxCeleb2 \cite{chung18b_interspeech}.}
  \label{fig:figure_same}
\end{figure*}

\begin{figure*}[t]
  \centering
  \includegraphics[width=0.99\columnwidth]{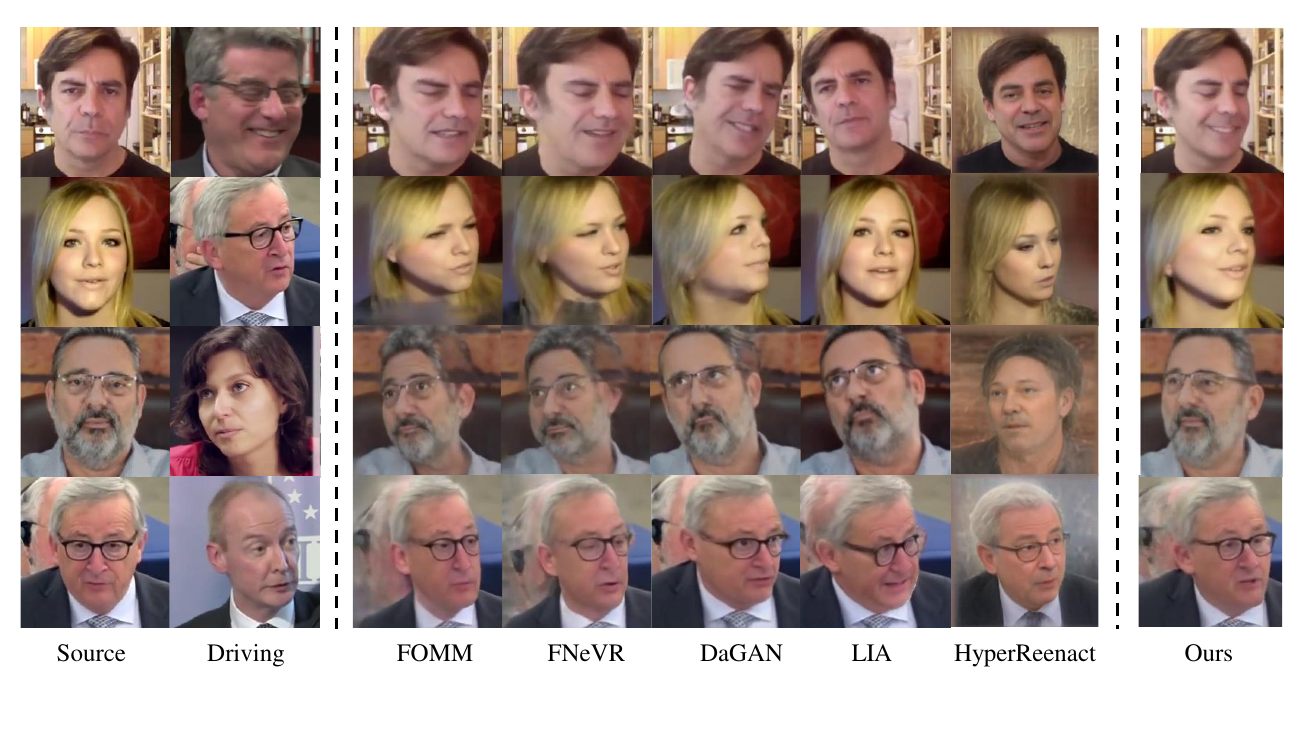}
  \caption{Cross-identity reenactment: demonstrating our method's superiority in geometry reconstruction and photorealistic face generation through a Qualitative Comparison on the TK Dataset\cite{wang2021one}.}
  \label{fig:figure_cross}
\end{figure*}

\subsection{Comparison with State-of-the-Art}
We conducted a comparative analysis of G3FA against six prominent models: First Order Motion Model (FOMM) \cite{siarohin2019first}, One-Shot Free-View Neural Talking-Head Synthesis for Video Conferencing (face vid2vid) \cite{wang2021one}, Depth-aware Generative Adversarial Network for Talking Head Video Generation (DaGAN) \cite{hong2022depth}, Neural Volume Rendering for Face Animation (FNeVR) \cite{zeng2022fnevr}, Latent Image Animator (LIA) \cite{wang2022latent} and One-Shot Reenactment via Jointly Learning to Refine and Retarget Faces (HyperReenact) \cite{bounareli2023hyperreenact} on the VoxCeleb2 dataset \cite{chung18b_interspeech}. We used the official implementations of the models, except for face vid2vid, where we relied on an unofficial yet well-known implementation.

\subsection{Same-identity Reconstruction}
Same-identity reconstruction involves reconstructing the face of a single individual using both the source image and driving frames. In \cref{Table:comparison1}, we show the results of our method compared to recent state-of-the-art approaches. As demonstrated, our framework exhibits substantial improvements across the majority of metrics, yielding highly refined samples. Notably, even without incorporating the face reconstruction component, our model, when coupled with the FNeVR architecture-based talking head model, achieves superior results. \cref{fig:figure_same} shows a qualitative evaluation of our method for this scenario on two examples. Our method produces the most realistic images, providing the closest resemblance to the driving faces.

\subsection{Cross-identity Reenactment}
In this section, we compare the performance of our method with state-of-the-art approaches on both the VoxCeleb2 \cite{chung18b_interspeech} and TK \cite{wang2021one} datasets. In this scenario, we evaluate the generalization of our face animation model by utilizing two distinct identities for the source and driving frames. As indicated in the \cref{Table:comparison2}, our method attains the highest CSIM value compared to all other approaches which shows the better identity preservation of our work. In addition, our G3FA achieves the highest FID score among the majority, closely approaching LIA. 
This work leverages a pretrained StyleGAN2 model, enhancing FID values. However, qualitative evaluation reveals shortcomings in accurately mimicking head rotation, a factor not assessed in FID calculation due to the predominance of frontal face samples. Notably, as depicted in \Cref{fig:figure_cross}, our model consistently produces geometrically consistent results in extreme head poses compared to other approaches.

 
\section{Conclusion}
We presented a novel integration of 3D information derived from neural inverse rendering into an adversarial learning framework, employing an ensemble of discriminators for one shot talking head synthesis. Our framework takes advantage of the intrinsic characteristics of 3D geometry to enhance the synthesis process outperforming current state-of-the-art face reenactment models. Importantly, our method can easily be integrated with existing face animation architectures based on Generative Adversarial Networks (GANs), without requiring any modifications to their fundamental structure. Leveraging off-the-shelf geometry extraction modules and discriminators to provide feedback to the generator preserves inference speed while enhancing the quality of generated samples.
\section{Acknowledgments}
This research has been partially funded by the EU project CORTEX² (GA: Nr 101070192).
\bibliography{egbib}
\end{document}